\documentclass{midl} 

\usepackage{amsfonts} 
\usepackage{multicol}
\usepackage{amssymb}
\usepackage{cases}
\usepackage{overpic}
\usepackage{multirow}
\usepackage{soul}
\usepackage{enumitem}
\usepackage{graphicx}
\usepackage{rotating}
\usepackage{booktabs}
\usepackage{amsmath}
\usepackage{datetime2}
\usepackage{makecell}
\usepackage{hyperref}
\usepackage{pifont}

\def\0{{\bf 0}}
\def\1{{\bf 1}}

\usepackage{flushend}

\title[LymphomaMIL]{A Multicenter Benchmark of Multiple Instance Learning Models for Lymphoma Subtyping from HE-stained Whole Slide Images
}

\midlauthor{\Name{Rao Muhammad Umer \nametag{$^{1}$}}  \Email{umer.rao@helmholtz-munich.de}\\
\Name{Daniel Sens \nametag{$^{1}$}} \Email{daniel.sens@helmholtz-munich.de}\\
\Name{Jonathan Noll \nametag{$^{1}$}} \Email{jonathan.noll.leon@gmail.com}\\
\Name{Sohom Dey \nametag{$^{1}$}} \Email{sohom21d@gmail.com}\\
\Name{Christian Matek \nametag{$^{1,2,7}$}} \Email{christian.matek@uk-erlangen.de}\\
\Name{Lukas Wolfseher \nametag{$^{8}$}} \Email{lukas.wolfseher@stud.uni-regensburg.de}\\
\Name{Rainer Spang \nametag{$^{8}$}} \Email{Rainer.Spang@med.uni-regensburg.de}\\
\Name{Ralf Huss \nametag{$^{9}$}} \Email{huss@bio-m.org}\\
\Name{Johannes Raffler \nametag{$^{9}$}} \Email{Johannes.Raffler@uk-augsburg.de}\\
\Name{Sarah Reinke \nametag{$^{10}$}} \Email{sreinke@path.uni-kiel.de}\\
\Name{Ario Sadafi \nametag{$^{1,6}$}} \Email{ario.sadafi@helmholtz-munich.de}\\
\Name{Wolfram Klapper \nametag{$^{10}$}} \Email{wklapper@path.uni-kiel.de}\\
\Name{Katja Steiger \nametag{$^{6}$}} \Email{katja.steiger@tum.de}\\
\Name{Kristina Schwamborn \nametag{$^{6}$}} \Email{kschwamborn@tum.de}\\
\Name{Carsten Marr \nametag{$^{1,2,3,4,5}$}} \Email{carsten.marr@helmholtz-munich.de}\\
\\
\addr $^{1}$ Institute of AI for Health, Helmholtz Munich, Munich, Germany \\
\addr $^{2}$ Department of Medicine III, Ludwig-Maximilian-University Hospital, Munich, Germany \\
\addr $^{3}$ Computational Health Center \& Helmholtz AI, Helmholtz Munich, Neuherberg, Germany \\
\addr $^{4}$ German Cancer Consortium (DKTK), Partner Site Munich, Germany \\
\addr $^{5}$ Munich Center for Machine Learning (MCML), Munich, Germany \\
\addr $^{6}$ Technical University of Munich, Munich, Germany \\
\addr $^{7}$ Institute of Pathology, Erlangen, Germany \\
\addr $^{8}$ University of Regensburg, Regensburg, Germany \\
\addr $^{9}$ Institute for Digital Medicine, University Hospital, Augsburg, Germany \\
\addr $^{10}$ Institute of Pathology, University Hospital, Kiel, Germany \\
}

\begin{document}

\maketitle

\begin{abstract}
Timely and accurate lymphoma diagnosis is essential for guiding cancer treatment. Standard diagnostic practice combines hematoxylin and eosin (HE)-stained whole slide images with immunohistochemistry, flow cytometry, and molecular genetic tests to determine lymphoma subtypes, a process requiring costly equipment, and skilled personnel, causing treatment delays. Deep learning methods could assist pathologists by extracting diagnostic information from routinely available HE-stained slides directly, yet comprehensive benchmarks for lymphoma subtyping on multicenter data are lacking. In this work, we present the first multicenter lymphoma benchmark, covering four common lymphoma subtypes and healthy control tissue. We systematically evaluate five publicly available pathology foundation models (H-optimus-1, H0-mini, Virchow2, UNI2, Titan) combined with attention-based (AB-MIL) and transformer-based (TransMIL) multiple instance learning aggregators across three magnifications ($10\times$, $20\times$, $40\times$). On in-distribution test sets, models achieve multiclass balanced accuracies exceeding $80\%$ across all magnifications, with foundation models performing similarly, and aggregation methods showing comparable results. The magnification study reveals that $40\times$ resolution is sufficient, with no performance gains from higher resolutions or cross-magnification aggregation. However, on out-of-distribution test sets, performance drops substantially to around $60\%$, highlighting significant generalization challenges. To advance the field, larger multicenter studies covering additional rare lymphoma subtypes are needed. We provide an automated benchmarking pipeline to facilitate such future research. 
Our paper codes is publicly available at \url{https://github.com/RaoUmer/LymphomaMIL}.
\end{abstract}


\section{Introduction}
\label{sec:intro} 
 Cancer is one of the deadliest diseases and remains an insurmountable obstacle to advance the quality and expectancy of patients' life~\cite{bray2021ever}. Lymphoma is a blood cancer that originates in the lymphatic system, which is a critical part of human body’s immune system. It specifically arises from lymphocytes, white blood cells that play a key role in defending the body against infections. Lymphomas are broadly classified into two categories~\cite{khoury20225th}: Hodgkin lymphoma (HL) and non-Hodgkin lymphoma (NHL), with each category having numerous subtypes. The diagnosis of lymphoma involves a combination of clinical evaluation, medical imaging, and most importantly, biopsy of the affected tissue. The biopsy is examined under a microscope (i.e., digitized as gigapixel Hematoxylin and Eosin (HE) stained whole slide images (WSIs)), and additional tests like immunohistochemical (IHC) stains, flow cytometry, cytogenetic, and molecular analysis help to determine the specific subtype of lymphoma~\cite{khoury20225th}. These auxiliary tests require costly equipment, expensive reagents, and trained personnel. Treatment varies depending on the lymphoma subtype, stage, and other factors such as the patient’s overall health. Common treatment options~\cite{khoury20225th} include chemotherapy, radiation therapy, targeted therapy, immunotherapy, and stem cell transplantation. 

Histopathology plays a central role in tissue-based diagnostics as a basis for understanding cellular and molecular mechanisms of disease by microscopic evaluation of morphological changes~\cite{campanella2023computational}.  Advances in histology slide scanning enable the digitization of histology slides into a whole slide image, paving the way for computer-aided diagnosis. In contrast to IHC, flow cytometry, and molecular genetic tests, HE-stained slides are inexpensive and widely available.  Recent studies~\cite{vorontsov2024foundation-virchow1,zimmermann2024virchow2,chen2024towards,hoptimus1,ding2025titan,h0-mini} have shown rapid progress in pathology foundation models (FMs) on diverse large unlabeled WSI datasets. These models learn transferable representations that can be adapted to a wide range of downstream pathology tasks. Among the most prominent pathology FMs are Virchow2~\cite{zimmermann2024virchow2}, UNI2~\cite{chen2024towards}, H-optimus-1~\cite{hoptimus1}, Titan~\cite{ding2025titan}, and H0-mini~\cite{h0-mini}. For downstream pathology tasks,  Multiple Instance Learning (MIL) has been applied as a key methodology in computational pathology \cite{ ilse2018attention,campanella2019clinical,sadafi2020attention,shao2021transmil}, where labels are assigned at the “bag” (WSI) level, while learning is performed from individual instances (patches) within each bag. MIL is particularly well-suited for histopathology, where detailed annotations are often difficult to obtain.
\begin{figure}[t!]
\centering
\includegraphics[width=1.0\textwidth]{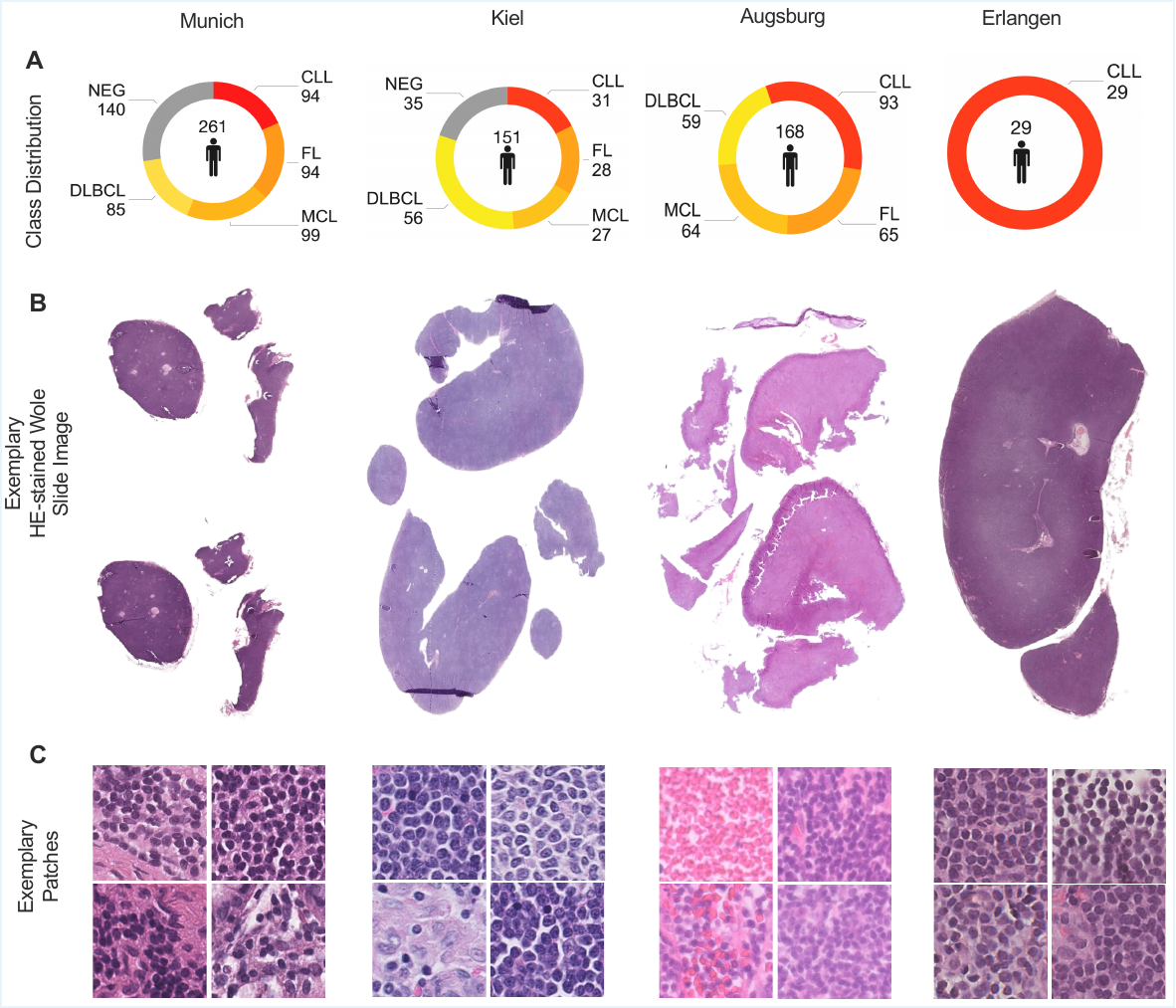}
\caption{A: Our multicenter lymphoma benchmark datasets comprises 5 classes : Follicular Lymphoma (FL), Chronic Lymphocytic Leukemia (CLL), Mantle Cell Lymphoma (MCL), Diffuse Large B-Cell Lymphoma (DLBCL) and healthy controls (NEG), with total $999$ HE WSIs from  $609$ individual patients. The total number of WSIs in each class are given. B: Thumbnails of exemplary CLL WSIs. C: Random patches of $256\times256$ pixels at $40\times$ magnification.}
\label{fig:ds_overview}
\end{figure}

\begin{figure}[t!]
\centering
\includegraphics[width=1.0\textwidth]{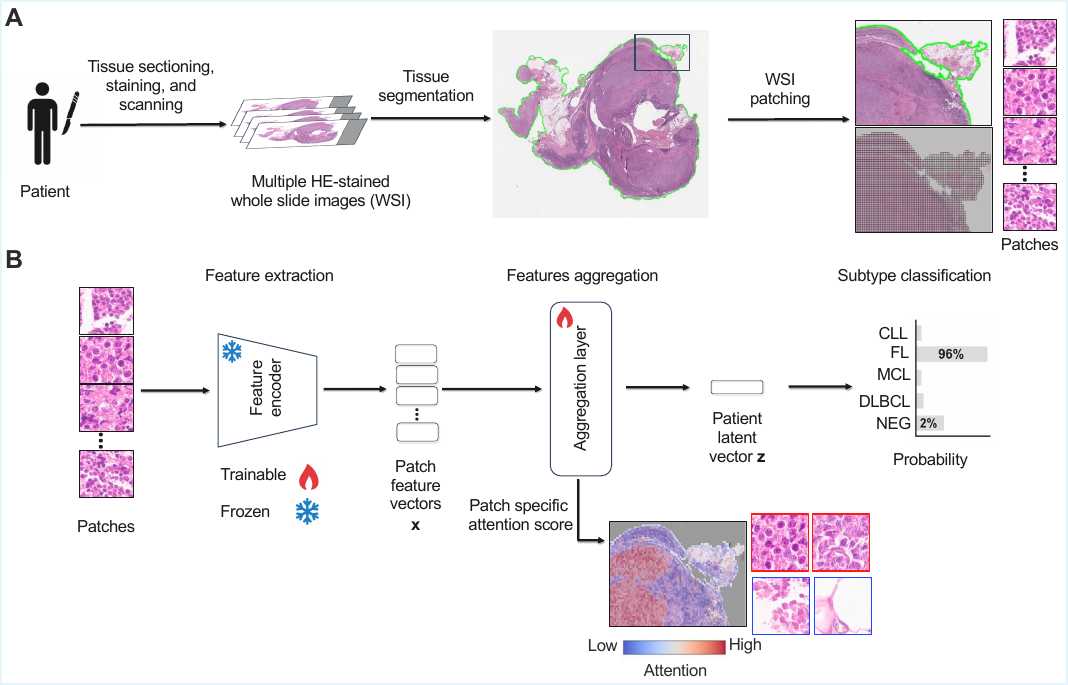}
\caption{A deep multiple instance learning (MIL) pipeline from whole slide images (WSI) acquisition to interpretability. A: Multiple WSIs are obtained per patient after tissue sectioning, HE staining, and digital scanning. Thousand of the patches are extracted from each WSI after segmenting the tissue regions. B: Extracted patches are passed to a pretrained encoder model (frozen) to extract feature vectors. The stacked feature vectors are fed to an attention based aggregation layer (trainable) to aggregate patch-level information into slide-level representations, which are used to make the final lymphoma subtyping prediction. For each subtype, we extract attention scores to visualize important regions of the WSI.}
\label{fig:method}
\end{figure}

Earlier studies~\cite{vrabac2021dlbcl,2010autocls,vorontsov2024foundation-virchow1} have demonstrated strong performance in distinguishing a limited set of lymphoma types, such as Diffuse Large B-Cell Lymphoma (DLBCL) and Follicular Lymphoma (FL) from HE-stained WSI. Their clinical utility remains limited due to narrow subtype coverage. In this work, we present the first multicenter lymphoma benchmark covering four common subtypes as well as healthy control tissue (Figure \ref{fig:ds_overview}). Moreover, we provide an automated benchmarking MIL pipeline for external evaluation (Figure~\ref{fig:method}). Our main contributions are the following:

\begin{itemize}
    \item We present a first multicenter lymphoma benchmark for the four most frequent subtypes and healthy controls. 
    \item We provide a deep MIL pipeline with five publicly available pathology foundation models for benchmarking.
    \item We evaluate FMs with attention-based MIL on in-distribution and out-of-distribution lymphoma datasets at three different resolutions.
    
\end{itemize}

\section{Methodology}
The deep learning pipeline for our experiments consists of lymphoma WSI, tissue segmentation and patch extraction, feature extraction, and lymphoma subtype classification (Figure \ref{fig:method}). In the following, we describe each component in detail.

\subsection{Lymphoma subtypes}
The lymphoma subtypes in the datasets (Figure \ref{fig:ds_overview}) are Chronic Lymphocytic Leukemia (CLL), Follicular Lymphoma (FL), Mantle Cell Lymphoma (MCL), and Diﬀuse Large B-Cell Lymphoma (DLBCL). In addition, the dataset contains WSIs showing only healthy tissue sections, which are used as a control class and designated as negative (NEG). CLL is a B-cell neoplasm that originates in the bone marrow and is characterized by the progressive accumulation of mature-appearing lymphocytes, which expand into the peripheral blood and infiltrate lymphoid organs, ultimately impairing normal organ function \cite{ghia2007chronic}. FL arises from germinal-center B-cells—centrocytes and centroblasts that proliferate aberrantly and form characteristic follicular or follicle-like structures \cite{xerri2016heterogeneity}. MCL develops from mantle-zone B-cells and typically presents with generalized lymphadenopathy or splenomegaly; it is defined by the expansion of cyclin-D1–driven malignant lymphocytes and displays an aggressive clinical course \cite{schieber2018current}. DLBCL represents a high-grade B-cell lymphoma in which antibody-producing lymphocytes undergo malignant transformation, leading to rapidly enlarging masses in lymph nodes or extranodal sites due to uncontrolled proliferation \cite{morton2006lymphoma}.

\begin{figure}[t!]
\centering
\includegraphics[width=0.8\textwidth]{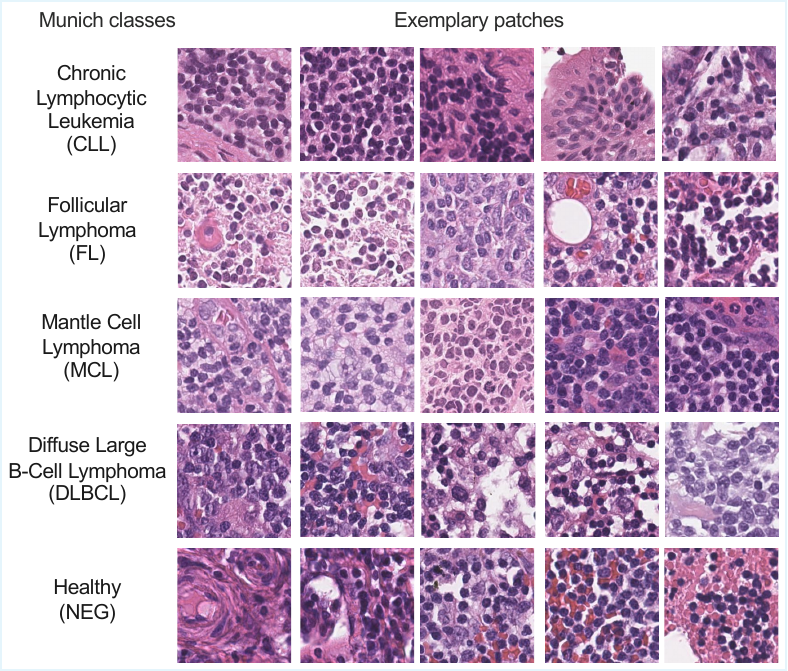}
\caption{Exemplary patches from five lymphoma classes in the Munich dataset. Each patch has a size of $256\times256$ pixels with $40\times$ magnification, at which individual cells become recognisable.}
\label{fig:cls_overview}
\end{figure}
\subsubsection{Munich}
Histology slides have been selected from the archive of the Institute of Pathology, Technical University of  Munich, Germany. The dataset consists of 513 HE-stained WSIs obtained from 261 patients containing tissue segments from lymph node biopsies. The WSIs were scanned at 40$\times$ magnification with Aperio AT2 scanner. 
The subtypes in the dataset are: FL, CLL, MCL, DLBCL, and NEG. Figure \ref{fig:cls_overview} shows the morphology of the five classes in dataset.
\subsubsection{Kiel}
Histology slides were selected from the archive of the Institute of Pathology, University Hospital, Kiel, Germany. The dataset consists of 177 HE-stained WSIs obtained from lymphoid tissue specimens of 151 patients. The WSIs were scanned at 40$\times$ magnification with a Hamamatsu scanner. It includes FL, CLL, MCL, DLBCL, and NEG. Some whole slides in the dataset originate from the same patient, as these represent the same biopsy scanned twice rather than a different section. Although the scans may exhibit slight color variations, their overall appearance remains largely similar (Figure \ref{fig:ds_overview}).
\subsubsection{Augsburg}
Histology slides were selected from the archive of the University Hospital, Augsburg, Germany. The dataset consists of de-identified 290 HE-stained WSIs obtained from 168 patients. The WSIs were scanned at 40$\times$ magnification with a Philips scanner. The subtypes in the dataset are FL, CLL, MCL, and DLBCL (Figure \ref{fig:ds_overview}).  
\subsubsection{Erlangen}
Histology slides were selected from the archive of the Institute of Pathology, University Hospital, Erlangen, Germany. The dataset consists of 29 HE-stained WSIs from 29 patients. The WSIs were scanned at 40$\times$ magnification. It includes only  the CLL class (Figure \ref{fig:ds_overview}).

\subsection{Tissue segmentation and patch extraction}
Tissue segmentation filters out background areas in WSIs, reducing the number of irrelevant patches and speeding up patch extraction. Traditional techniques rely on image-processing heuristics such as binary or Otsu thresholding, which require manual parameter tuning and often fail to generalize across stains. These approaches also struggle to distinguish true tissue from common artifacts, including pen marks and bubbles. Trident~\cite{zhang2025standardizing} is a recent alternative for robust and stain-agnostic tissue segmentation and patch extraction. After tissue segmentation, the patching step divides the tissue-containing regions into $N$ non-overlapping image patches with a size of $256 \times 256$ pixel at specific magnification for later processing by a feature encoder (Figure \ref{fig:method}A). We can represent the resulting patches as $\mathbf{X}=\left\{\mathbf{x}_1, \ldots, \mathbf{x}_N\right\}, \mathbf{x}_i \in[0,255]^{256 \times 256 \times 3}$, with the i-th patch.

\subsection{Feature extraction}
Due to the high dimensionality of a WSI, $\mathbf{X}$ usually consists of a large number of instances and cannot be processed by the downstream aggregation models. Thus, we first pass patches to a pretrained encoder to yield patch features by compressing the high dimensional visual information into a one dimensional feature vector. Lets suppose $f_\psi$ is a pretrained encoder with parameters $\psi$. For each patch $\mathbf{x}_i \in \mathbf{X}$ the corresponding feature vector is obtained as $\mathbf{h}_i = f_\psi\left(\mathbf{x}_i\right)$. The WSI is the represented by a bag of instance features $\mathbf{H}=\left\{\mathbf{h}_1, \ldots, \mathbf{h}_N\right\}$, which is passed as an input to the aggregation layer (Figure~\ref{fig:method}B). In our pipeline, the pretrained encoders can either be used to extract patch-level or slide-level features.

\subsection{Lymphoma subtype classification}
We use the deep multiple instance learning (MIL) framework for the trainable aggregation layer to classify the bag $\mathbf{H}$ of instance features into the five classes of our lymphoma dataset. The aggregation layer takes as input a bag $\mathbf{H}$ of $N$ instance features and outputs a patient latent vector $\mathbf{z}=\sum_{k=1}^N a_k \mathbf{h}_k$, where $a_k$ is the attention score. 
Finally, the patient latent vector $\mathbf{z}$ is passed to classifier to obtain predicted class probabilities. During training, the cross-entropy loss is computed between the predicted WSI class label $\mathbf{\hat{Y}}$ from the probability and ground-truth label $\mathbf{Y} \in \left\{1, \ldots, C\right\}$.  
\begin{table}[t!]
    \centering
    \caption{\textbf{Magnification study}: Five-fold cross-validation results for lymphoma subtype classification on Munich dataset at different magnification level, $40\times$, $20\times$, and $10\times$. Performance is reported in terms of Area Under ROC (AUC), Macro F1-score, and balanced accuracy (BACC).}
    \tabcolsep=0.01\linewidth
    \resizebox{1.0\textwidth}{!}{
    \begin{tabular}{ccccccc}
    \toprule
     \textbf{Magnification} & \textbf{Encoder} & \makecell{\textbf{Embedding} \\ \textbf{Dim}} & \textbf{Aggregator}  & \textbf{AUC} & \textbf{F1-score} & \textbf{BACC} \\
    \midrule
      &  \multirow{3}{*}{\makecell{ResNet50 \\ (Beseline)}}  & \multirow{3}{*}{1024} & AB-MIL &  0.89 \scriptsize$\pm$0.03 & 0.66 \scriptsize$\pm$0.06 & 0.67 \scriptsize$\pm$0.06 \\
      & &  & TransMIL & 0.85 \scriptsize$\pm$0.02 & 0.58 \scriptsize$\pm$0.02 & 0.59 \scriptsize$\pm$0.03 \\
      &   &  & TransMIL + BEL & 0.87 \scriptsize$\pm$0.02 & 0.62 \scriptsize$\pm$0.04 & 0.63 \scriptsize$\pm$0.03 \\
      \cline{2-7}
      &  \multirow{3}{*}{\makecell{UNI2 \\ (Patch-level)}} & \multirow{3}{*}{1536} & AB-MIL &  0.95 \scriptsize$\pm$0.01 & 0.79 \scriptsize$\pm$0.02 & 0.79 \scriptsize$\pm$0.03 \\
    $40 \times$ &  &  & TransMIL & 0.94 \scriptsize$\pm$0.01 & 0.75 \scriptsize$\pm$0.04 & 0.75 \scriptsize$\pm$0.04 \\
      &   &  & TransMIL + BEL & \textbf{0.96} \scriptsize$\pm$0.01 & 0.80 \scriptsize$\pm$0.03 & 0.80 \scriptsize$\pm$0.03 \\
      \cline{2-7}
      & \multirow{4}{*}{\makecell{Titan \\ (Slide-level)}}  & \multirow{4}{*}{768} & Linear &  0.95 \scriptsize$\pm$0.01 & 0.78 \scriptsize$\pm$0.02 & 0.78 \scriptsize$\pm$0.01 \\
      & & & AB-MIL &  \textbf{0.96} \scriptsize$\pm$0.01 & \textbf{0.83} \scriptsize$\pm$0.06 & \textbf{0.82} \scriptsize$\pm$0.05 \\
      &  &  & TransMIL & \textbf{0.96} \scriptsize$\pm$0.01 & 0.82 \scriptsize$\pm$0.03 & \textbf{0.82} \scriptsize$\pm$0.03 \\
      &   &   & TransMIL + BEL & \textbf{0.96} \scriptsize$\pm$0.01 & 0.82 \scriptsize$\pm$0.04 & 0.81 \scriptsize$\pm$0.04 \\
      \midrule

     &  \multirow{3}{*}{\makecell{ResNet50 \\ (Beseline)}}  & \multirow{3}{*}{1024} & AB-MIL &  0.89 \scriptsize$\pm$0.02 & 0.63 \scriptsize$\pm$0.04 & 0.64 \scriptsize$\pm$0.04 \\
      & &  & TransMIL & 0.81 \scriptsize$\pm$0.04 & 0.50 \scriptsize$\pm$0.04 & 0.51 \scriptsize$\pm$0.04 \\
      &   &  & TransMIL + BEL & 0.84 \scriptsize$\pm$0.02 & 0.58 \scriptsize$\pm$0.02 & 0.58 \scriptsize$\pm$0.03 \\
      \cline{2-7}
      &  \multirow{3}{*}{\makecell{UNI2 \\ (Patch-level)}} & \multirow{3}{*}{1536} & AB-MIL &  0.95 \scriptsize$\pm$0.02 & 0.77 \scriptsize$\pm$0.03 & 0.77 \scriptsize$\pm$0.03 \\
    $20 \times$  &  &  & TransMIL & 0.94 \scriptsize$\pm$0.01 & 0.79 \scriptsize$\pm$0.04 & 0.79 \scriptsize$\pm$0.04 \\
      &   &  & TransMIL + BEL & \textbf{0.96} \scriptsize$\pm$0.01 & \textbf{0.83} \scriptsize$\pm$0.03 & \textbf{0.83} \scriptsize$\pm$0.03 \\
      \cline{2-7}
      & \multirow{4}{*}{\makecell{Titan \\ (Slide-level)}}  & \multirow{4}{*}{768} & Linear &  0.95 \scriptsize$\pm$0.01 & 0.77 \scriptsize$\pm$0.02 & 0.77 \scriptsize$\pm$0.02 \\
      & & & AB-MIL &  \textbf{0.96} \scriptsize$\pm$0.01 & 0.82 \scriptsize$\pm$0.06 & 0.81 \scriptsize$\pm$0.07 \\
      &  &  & TransMIL & \textbf{0.96} \scriptsize$\pm$0.01 & 0.82 \scriptsize$\pm$0.04 & 0.82 \scriptsize$\pm$0.04 \\
      &   &   & TransMIL + BEL & \textbf{0.96} \scriptsize$\pm$0.01 & 0.82 \scriptsize$\pm$0.03 & 0.82 \scriptsize$\pm$0.04 \\
      \midrule

       &  \multirow{3}{*}{\makecell{ResNet50 \\ (Beseline)}}  & \multirow{3}{*}{1024} & AB-MIL & 0.85 \scriptsize$\pm$0.02 & 0.60 \scriptsize$\pm$0.07 & 0.61 \scriptsize$\pm$0.06 \\
      & &  & TransMIL & 0.83 \scriptsize$\pm$0.04 & 0.52 \scriptsize$\pm$0.06 & 0.54 \scriptsize$\pm$0.06 \\
      &   &  & TransMIL + BEL & 0.82 \scriptsize$\pm$0.04 & 0.51 \scriptsize$\pm$0.04 & 0.53 \scriptsize$\pm$0.04 \\
      \cline{2-7}
      &  \multirow{3}{*}{\makecell{UNI2 \\ (Patch-level)}} & \multirow{3}{*}{1536} & AB-MIL &  0.95 \scriptsize$\pm$0.01 & 0.80 \scriptsize$\pm$0.04 & 0.80 \scriptsize$\pm$0.04 \\
    $10 \times$  &  &  & TransMIL & 0.95 \scriptsize$\pm$0.01 & 0.80 \scriptsize$\pm$0.04 & 0.80 \scriptsize$\pm$0.04 \\
      &   &  & TransMIL + BEL & \textbf{0.96} \scriptsize$\pm$0.02 & \textbf{0.81} \scriptsize$\pm$0.05 & \textbf{0.81} \scriptsize$\pm$0.05 \\
      \cline{2-7}
      & \multirow{4}{*}{\makecell{Titan \\ (Slide-level)}}  & \multirow{4}{*}{768} & Linear &  0.94 \scriptsize$\pm$0.01 & 0.75 \scriptsize$\pm$0.04 & 0.75 \scriptsize$\pm$0.04 \\
      & & & AB-MIL &  0.95 \scriptsize$\pm$0.02 & 0.79 \scriptsize$\pm$0.04 & 0.79 \scriptsize$\pm$0.04 \\
      &  &  & TransMIL & 0.94 \scriptsize$\pm$0.01 & 0.77 \scriptsize$\pm$0.04 & 0.78 \scriptsize$\pm$0.04 \\
      &   &   & TransMIL + BEL & 0.95 \scriptsize$\pm$0.01 & 0.80 \scriptsize$\pm$0.04 & 0.80 \scriptsize$\pm$0.04 \\

      \midrule

      $40 \times$ , $20 \times$  , $10 \times$  &  \makecell{UNI2 \\ (Patch-level)} & 1536 & MoA &  0.94 \scriptsize$\pm$0.03 & 0.76 \scriptsize$\pm$0.08 & 0.77 \scriptsize$\pm$0.07 \\
    \bottomrule
    \end{tabular}}
    \label{tab:results-diff-mag}
\end{table}

\begin{table}[htb]
    \centering
    \caption{\textbf{In-domain comparison of publicly available feature encoders}: Five-fold cross-validation results for lymphoma subtype classification on Munich and Kiel lymphoma datasets at 40x magnification. Performance is reported in terms of AUC, F1-score, and balanced accuracy (BACC).}
    \tabcolsep=0.01\linewidth
    \resizebox{0.9\textwidth}{!}{
    \begin{tabular}{lcccccc}
    \toprule
    & & & & \multicolumn{3}{c}{\textbf{IID Test}}\\
    \cmidrule(lr){5-7}
     \textbf{Dataset} & \textbf{Encoder} & \makecell{\textbf{Embedding} \\ \textbf{Dim}} & \textbf{Aggregator}  & \textbf{AUC} & \textbf{F1-score} & \textbf{BACC} \\
    \midrule
      \multirow{6}{*}{Munich}  
       &  ResNet50  & 1024 & \multirow{6}{*}{AB-MIL} &  0.89 \scriptsize$\pm$0.03 & 0.66 \scriptsize$\pm$0.06 & 0.67 \scriptsize$\pm$0.06 \\
       &  H-optimus-1 & 1536 &  &  \textbf{0.96} \scriptsize$\pm$0.01 & 0.79 \scriptsize$\pm$0.03 & 0.79 \scriptsize$\pm$0.03 \\
       &  H0-mini & 1536 &  &  0.95 \scriptsize$\pm$0.01 & 0.79 \scriptsize$\pm$0.02 & 0.78 \scriptsize$\pm$0.01 \\
       &  Virchow2 & 2560 &  &  \textbf{0.96} \scriptsize$\pm$0.01 & 0.77 \scriptsize$\pm$0.04 & 0.77 \scriptsize$\pm$0.04 \\
      &  UNI2 & 1536 &  &  0.95 \scriptsize$\pm$0.01 & 0.79 \scriptsize$\pm$0.02 & 0.79 \scriptsize$\pm$0.03 \\
      & Titan  & 768 &  &  \textbf{0.96} \scriptsize$\pm$0.01 & \textbf{0.83} \scriptsize$\pm$0.06 & \textbf{0.82} \scriptsize$\pm$0.05 \\
    
    \midrule

    \multirow{6}{*}{Kiel}  
       &  ResNet50  & 1024 & \multirow{6}{*}{AB-MIL} &  0.85 \scriptsize$\pm$0.05 & 0.64 \scriptsize$\pm$0.08 & 0.65 \scriptsize$\pm$0.04 \\
       &  H-optimus-1 & 1536 &  &  \textbf{0.97} \scriptsize$\pm$0.01 & 0.81 \scriptsize$\pm$0.09 & 0.81 \scriptsize$\pm$0.09 \\
       &  H0-mini & 1536 &  &  0.92 \scriptsize$\pm$0.05 & 0.66 \scriptsize$\pm$0.15 & 0.69 \scriptsize$\pm$0.13 \\
       &  Virchow2 & 2560 &  &  0.96 \scriptsize$\pm$0.02 & 0.72 \scriptsize$\pm$0.11 & 0.74 \scriptsize$\pm$0.09 \\
      &  UNI2 & 1536 &  &  0.96 \scriptsize$\pm$0.03 & 0.82 \scriptsize$\pm$0.07 & \textbf{0.82} \scriptsize$\pm$0.06 \\
      & Titan  & 768 &  &  0.96 \scriptsize$\pm$0.02 & \textbf{0.83} \scriptsize$\pm$0.06 & 0.81 \scriptsize$\pm$0.05 \\

    \bottomrule
    \end{tabular}}
    \label{tab:results-diff-encoder}
\end{table}

\section{Experimental Results}
\subsection{Preprocessing}
Our HE-stained WSIs contain pen marks and artifacts, i.e., blurring, compression, water bubbles. We use Trident~\cite{zhang2025standardizing} for both tissue segmentation and patch extraction; it employs a DeepLabV3-based segmentation model that provides more robust and stain-agnostic tissue detection and segmentation, and further extract non-overlapping patches from a WSI.

\subsection{Evaluation}
We split the data patient-wise for five-fold cross-validation with train, validation, and test sets with a ratio of $80\%$, $10\%$, and $10\%$, to assess the performance of trained MIL models. To prevent data leakage and overfitting, all images of one patient are assigned to the same fold. The training set is shuffled to introduce slight variability, while the validation and test sets remain unshuffled to ensure consistent and reliable evaluation. 

The model’s performance is evaluated using the Area Under the ROC Curve (AUC), Macro F1-score, and Balanced Accuracy (BACC) metrics. These metrics are calculated on both the validation and test datasets to provide a comprehensive assessment of classification effectiveness. Additionally, confusion matrices are analyzed to interpret the distribution of prediction errors.

\subsection{Implementation details}
We use six pretrained vision encoders for feature extraction, ResNet50 (baseline) \cite{he2016resnet50}, H-optimus-1 \cite{hoptimus1}, H0-mini \cite{h0-mini}, Virchow2 \cite{zimmermann2024virchow2}, UNI2 \cite{chen2024uni}, Titan \cite{ding2025titan}. Each model independently process the WSI patches, generating d-dimensional embeddings encoding the visual representation of each WSI (Figure~\ref{fig:method}B).

The MIL models are trained under a weakly supervised learning setting, as only slide-level labels are available and no detailed annotations for individual patches. We use the MIL architecture, employing four variants: attention-based deep MIL (AB-MIL) \cite{ilse2018attention}, transformer-based MIL (TransMIL) \cite{shao2021transmil}, TransMIL + BEL \cite{sens2023bel}, and MoA~\cite{ozlugedik2026moa}. These models aggregate features extracted from individual instances (patches) within each WSI to generate a single, comprehensive representation of the entire slide. By learning to assign different weights to instances, the architectures can capture the most relevant morphological features contributing to the classification task. We address class imbalance by applying balanced sampling at the patient level during aggregator model training, rather than relying on slide-level label balancing. This strategy mitigates subtype imbalance effects and ensures that model performance is not biased by overrepresented classes.

The MIL models are trained for 50 epochs. Each training step involves a forward pass to generate predictions, computation of the cross-entropy loss by comparing predictions with ground-truth slide-level labels, and parameter updates via backpropagation using the AdamW optimizer, which combines weight decay with adaptive learning rate adjustments. To support convergence, a learning rate schedule with a warm-up phase of 10 epochs is used, linearly increasing the learning rate from 0 to 0.0001. After the warm-up, a cosine decay scheduler gradually reduces the learning rate down to 1e-6. Simultaneously, the weight decay, initially set to 0.04, are gradually increased up to 0.4. A momentum factor of 0.9 is also applied to accelerate optimization.

\begin{table}[htb]
    \centering
    \caption{\textbf{Out-of-distribution generalization}: Five-fold cross-validation results for lymphoma subtype classification on external datasets at $40 \times$ magnification.  UNI2 and Titan are used as feature encoders and pretrained aggregators (AB-MIL and TransMIL) on Munich dataset. Performance is reported in terms of AUC, F1-score, and balanced accuracy (BACC). }
    \tabcolsep=0.01\linewidth
    \resizebox{0.8\textwidth}{!}{
    \begin{tabular}{llcccc}
    \toprule
    & & & \multicolumn{3}{c}{\textbf{OOD Test}}\\
    \cmidrule(lr){4-6}
     \textbf{Encoder} & \makecell{\textbf{OOD} \\ \textbf{Dataset}} & \textbf{Aggregator}  & \textbf{AUC} & \textbf{F1-score} & \textbf{BACC} \\
    \midrule
      \multirow{6}{*}{UNI2}  
       &  \multirow{2}{*}{Kiel}  & AB-MIL & \textbf{0.88} \scriptsize$\pm$0.05 & \textbf{0.61} \scriptsize$\pm$0.06 & \textbf{0.62} \scriptsize$\pm$0.04  \\
       & & TransMIL & 0.86 \scriptsize$\pm$0.06 & 0.58 \scriptsize$\pm$0.06 & 0.60 \scriptsize$\pm$0.04 \\
       &  \multirow{2}{*}{Augsburg}  & AB-MIL & \textbf{0.82} \scriptsize$\pm$0.04 & \textbf{0.58} \scriptsize$\pm$0.05 & \textbf{0.60} \scriptsize$\pm$0.06  \\
       & & TransMIL & 0.77 \scriptsize$\pm$0.05 & 0.50 \scriptsize$\pm$0.05 & 0.51 \scriptsize$\pm$0.05  \\
       &  \multirow{2}{*}{Erlangen}  & AB-MIL & 0.75 \scriptsize$\pm$0.05 & 0.37 \scriptsize$\pm$0.07 & 0.59 \scriptsize$\pm$0.18  \\
       & & TransMIL & \textbf{0.90} \scriptsize$\pm$0.05 & \textbf{0.75} \scriptsize$\pm$0.35 & \textbf{0.83} \scriptsize$\pm$0.24  \\
    
    \midrule

    \multirow{6}{*}{Titan}  
       &  \multirow{2}{*}{Kiel}  & AB-MIL & \textbf{0.88} \scriptsize$\pm$0.05 & 0.43 \scriptsize$\pm$0.08 & 0.48 \scriptsize$\pm$0.08  \\
       & & TransMIL & 0.84 \scriptsize$\pm$0.04 & \textbf{0.45} \scriptsize$\pm$0.10 & \textbf{0.50} \scriptsize$\pm$0.08  \\
       &  \multirow{2}{*}{Augsburg}  & AB-MIL & \textbf{0.80} \scriptsize$\pm$0.03  &  \textbf{0.41} \scriptsize$\pm$0.06 & \textbf{0.48} \scriptsize$\pm$0.04  \\
       & & TransMIL & 0.77 \scriptsize$\pm$0.03  &  0.37 \scriptsize$\pm$0.06 & 0.46 \scriptsize$\pm$0.04  \\
       &  \multirow{2}{*}{Erlangen}  & AB-MIL & 0.78 \scriptsize$\pm$0.05 & 0.35 \scriptsize$\pm$0.09 & 0.53 \scriptsize$\pm$0.20  \\
       & & TransMIL & \textbf{0.86} \scriptsize$\pm$0.05 &  \textbf{0.60} \scriptsize$\pm$0.25 & \textbf{0.72} \scriptsize$\pm$0.18  \\

    \bottomrule
    \end{tabular}}
    \label{tab:results-ood}
\end{table}

\begin{figure}[t!]
\centering
\includegraphics[width=0.9\textwidth]{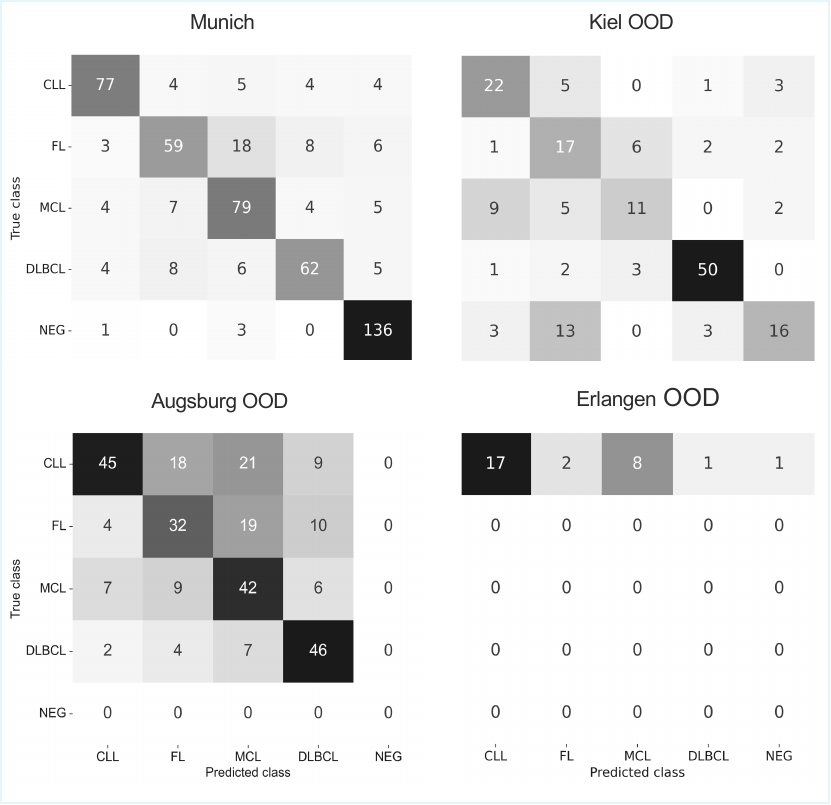}
\caption{Confusion matrix comparison for UNI + AB-MIL on in-distribution Munich and out-of-distribution testsets from Kiel, Augsburg, and Erlangen.}
\vspace{-0.3cm}
\label{fig:conf_matrices}
\end{figure}
\subsection{Classification results}

For lymphoma subtype prediction, we assessed model performance across varying magnifications, multiple patch-level and slide-level feature encoders, using both in-distribution and out-of-distribution (OOD) test sets. Our lymphoma subtyping findings should be interpreted as complementary to the “all-task” benchmark~\cite{bareja2025evaluating} rather than directly comparable, because the benchmark does not include lymphoma subtyping, and TCGA offers only a single lymphoma subtype (DLBCL) with limited histological diversity.

\begin{figure}[t!]
\centering
\includegraphics[width=0.95\textwidth]{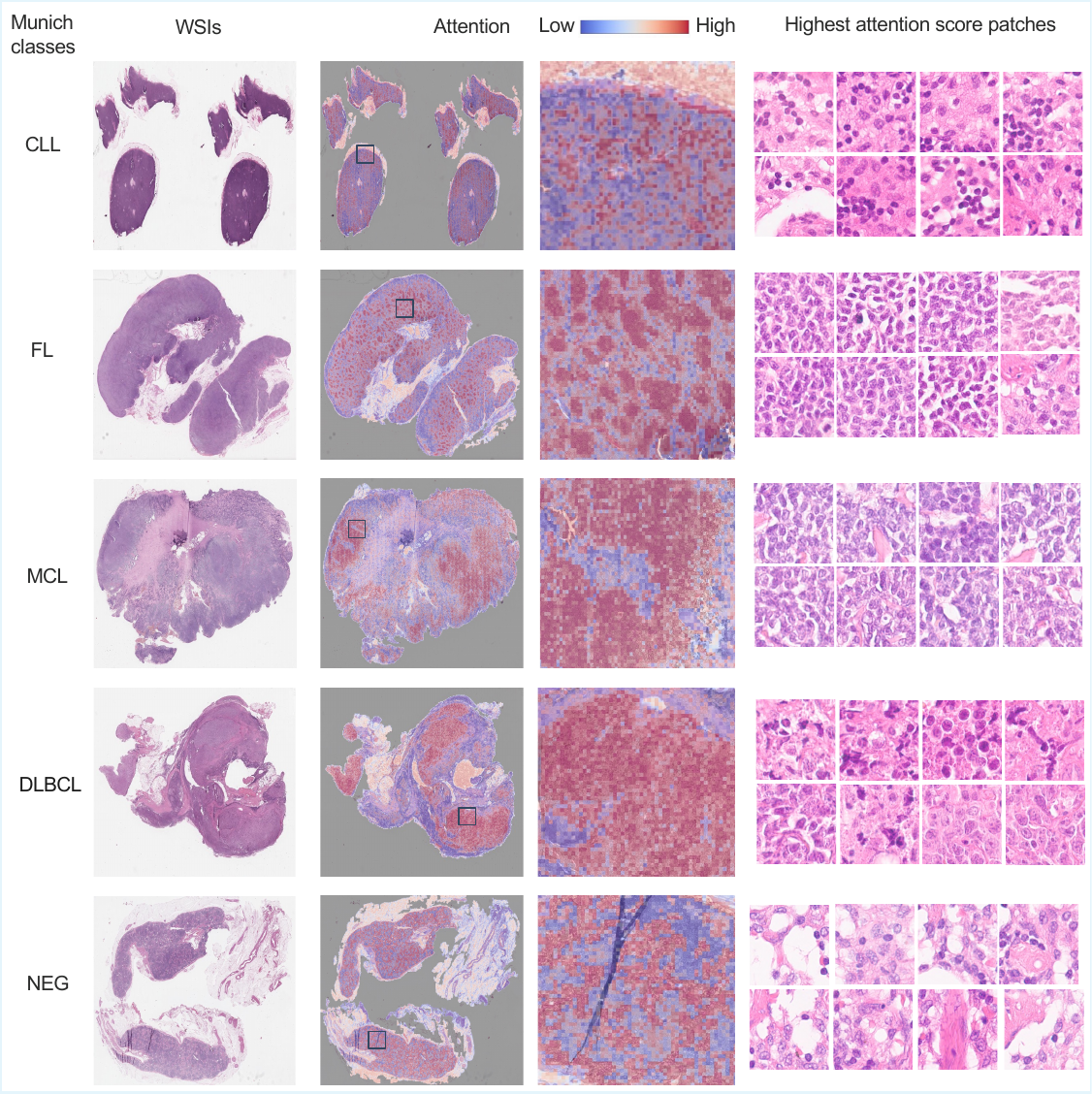}
\caption{Attention visualization on IID Munich testset. Highest attention regions correspond to morphology, while lower attention regions mostly correspond to normal tissue or artifacts patches.}
\label{fig:diff_classes_attn_comparison}
\end{figure}

\textbf{Algorithm combination and magnification study (Table~\ref{tab:results-diff-mag}):}
On the in-distribution Munich test set, we compared baseline ResNet50, the patch-level encoder UNI2, and the slide-level encoder Titan across 40×, 20×, and 10× magnifications. TransMIL combined with UNI2 or Titan is better than AB-MIL at 40× and 20×, with a slight performance decline observed at 10×. Despite this, AB-MIL remained the most computationally efficient option for training and inference compared to transformer-based MIL methods. TransMIL + BEL show only minimal gains over TransMIL. This behavior is likely due to a mismatch between the BEL objective and the morphological characteristics of lymphoma WSIs: While BEL is designed to regularize bag-level embeddings in transformer-based MIL, lymphoma subtyping relies on heterogeneous and spatially dispersed patterns rather than a single dominant bag embedding. We believe that TransMIL already captures the discriminative context in the slides, leaving limited room for additional gains from BEL. We trained TransMIL + BEL using the original hyperparameters proposed by \cite{sens2023bel}, but checked that further parameter tuning did not lead to improvements. We explored more recent MIL extension, including multi-resolution fusion strategy such as mixtures of aggregator (MoA); however, it did not outperform the standard AB-MIL or TransMIL baselines in our experiments. In addition, as shown in Figures \ref{fig:diff_classes_attn_comparison}, \ref{fig:diff_models_attn_comparison}, and \ref{fig:diff_dataset_attn_comparison} in supplementary material, attention distributions remain highly heterogeneous across whole slides, indicating that informative regions are not trivially separable. This suggests that more advanced and principled patch-selection or multi-scale modeling strategies are needed, which we leave for future work.

\textbf{Comparison of publicly available feature encoders (Table~\ref{tab:results-diff-encoder}):}
Using AB-MIL, we evaluated several open-source pathology feature encoders on the Munich and Kiel in-distribution cohorts. UNI2 and Titan achieved superior performance across both datasets, indicating their strong representation capacity for lymphoma subtyping.

\textbf{Out-of-distribution generalization (Table~\ref{tab:results-ood}):}
Models trained on the Munich dataset using UNI2 or Titan features were evaluated on three external cohorts: Kiel, Augsburg, and Erlangen. 
\begin{table}
  \centering
  \caption{\textbf{Computational efficiency} statistics of five pathology foundation models, including parameter count, FLOPs, and inference time per image or per slide. Throughput is computed on images per second for patch-level models, and slides per second for slide-level model (Titan).}
  \begin{tabular}{lccc}
    \toprule
    \textbf{Model} & \textbf{Params.} & \textbf{FLOPs} & \textbf{Throughput} \\
      & [M] $(\downarrow)$ & [G] $(\downarrow)$ & [img or slide/s] $(\uparrow)$ \\
    \midrule
    
    H-optimus-1 &  1,135 & 591.81 & 82.5 \scriptsize$\pm$4.3   \\
    H0-mini & 87 & 44.60 & \textbf{885.1} \scriptsize$\pm$7.1 \\
    Virchow2 &  632 & 329.11 & 134.3 \scriptsize$\pm$7.9\\
    UNI2 &  681 & 360.71 & 134.6 \scriptsize$\pm$7.6 \\
    \midrule
    Titan & \textbf{85} & \textbf{36.76} & \textbf{43.95} \scriptsize$\pm$2.3 \\
    \bottomrule
  \end{tabular}
    \label{tab:performance}
\end{table}

UNI2 paired with AB-MIL achieved the best overall performance on OOD Kiel. UNI2 is trained on large-scale, diverse histopathology datasets and preserves fine-grained cellular morphology, which is critical for lymphoma subtyping \cite{chen2024uni}. In contrast, Titan operates at the slide level by aggressively reducing token resolution, which improves efficiency but limits its ability to capture subtle, spatially localized lymphoma patterns under domain shift. As a result, UNI2 features generalize more robustly to unseen centers.  

UNI2 paired with AB-MIL achieved the best overall performance on OOD Augsburg. We use Augsburg data to assess model robustness under domain shift. To address the label-space mismatch, we report four-class metrics using the original predictions, considering the top-2 class for the those cases where NEG is the top-1 prediction, resulting in comparable performance of our models on the Augsburg dataset.

For the Erlangen dataset, which contains only CLL cases, UNI2 combined with TransMIL yielded the highest accuracy. In the Erlangen cohort, the high standard deviation observed is primarily due to its very small sample size (29 WSIs) and its single-class composition (CLL only). In this setting, balanced accuracy becomes highly sensitive to fold-level variations, leading to inflated variance. Additionally, Erlangen data were acquired using a different scanner, further amplifying instability. We report these results transparently to highlight the challenges of evaluating OOD performance on small, single-class cohorts, rather than as evidence of reliable generalization. Across all OOD evaluations, UNI2 + AB-MIL demonstrated the most robust and stable generalization behavior.

\textbf{Confusion matrix comparison on in-distribution and out-of-distribution testsets:}
Figure~\ref{fig:conf_matrices} presents confusion matrices for UNI2 + AB-MIL on the in-distribution Munich test set and the three OOD cohorts. The model performed obviously strongest on the IID Munich dataset, followed by solid results on Kiel, Augsburg, and Erlangen.

\begin{figure}[t!]
\centering
\includegraphics[width=0.95\textwidth]{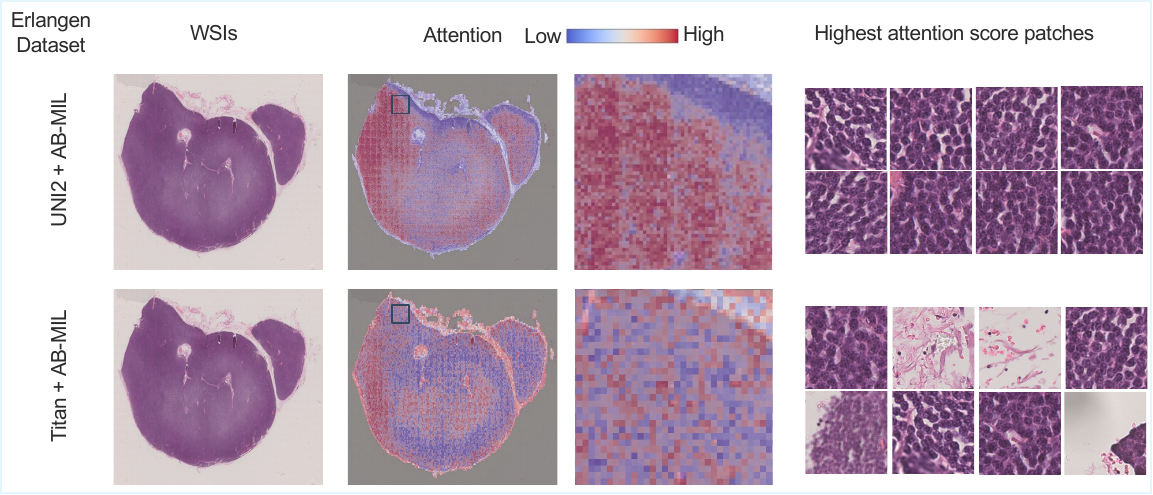}
\caption{Attention visualization of CLL lymphoma subtype on OOD Erlangen cohort. UNI2 avoids artifact patches in the highest attention scores as compared to Titan.}
\label{fig:diff_models_attn_comparison}
\end{figure}

\begin{figure}[t!]
\centering
\includegraphics[width=0.95\textwidth]{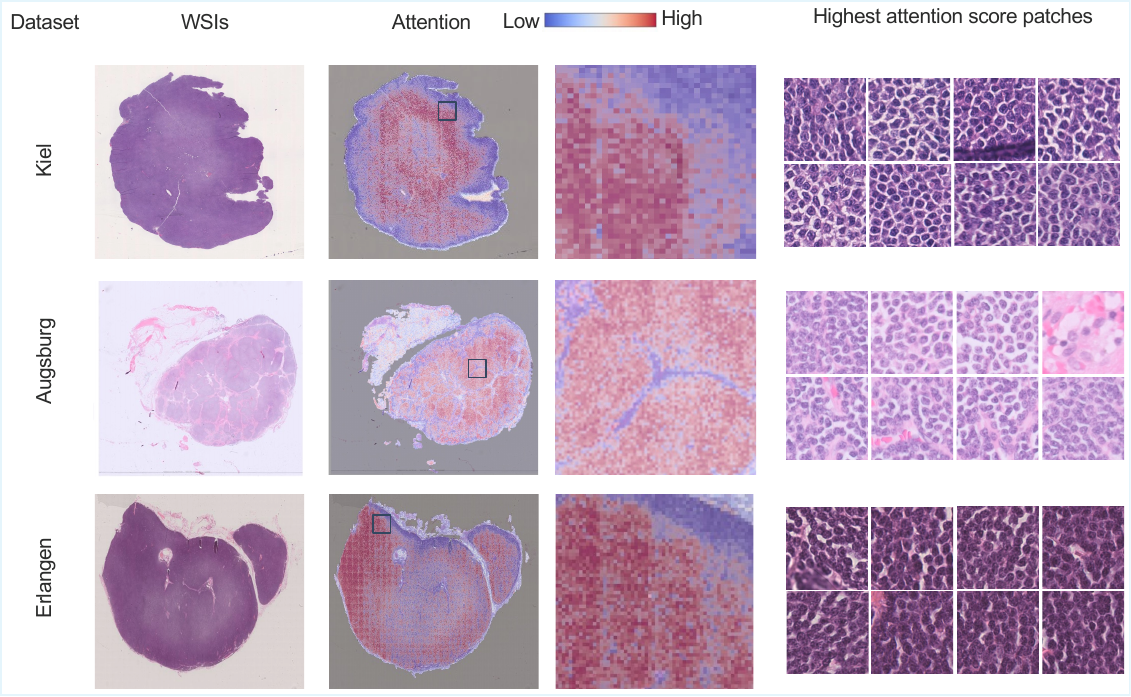}
\caption{Attention visualization of CLL lymphoma subtype on OOD lymphoma testsets.}
\label{fig:diff_dataset_attn_comparison}
\end{figure}

\textbf{Computational efficiency:}
Table~\ref{tab:performance} compares computational efficiency of the five pathology foundation models, reporting  parameters, FLOPs, and throughput (images per second) measured on a consumer-grade RTX 3090 GPU. For the patch-level FMs, throughput is computed in FP16 precision with a batch size of 32, and we report the mean and standard deviation over 500 batches. In WSI analysis, where thousands of image patches are encoded for a slide level prediction, this efficiency translates into total processing time with varying number of image patches per slide. For slide-level FMs such as Titan, it operates at the whole slide level by substantially reducing the number of processed tokens during inference. The throughput is computed as slides per second with 10,000 patches per slide on A100 GPU.

\section{Whole slide image attention visualization}
\label{sec:attn_viz}
The attention scores can be visualized as heatmaps to highlight diagnostically informative regions, areas assigned high attention, while de-emphasizing regions of low relevance, such as normal tissue or background artifacts. To interpret the spatial distribution of model attention across a WSI, we convert the attention scores corresponding to the predicted class into percentiles and map these normalized values back to their original coordinates on the slide. Fine-grained heatmaps are produced by extracting overlapping patches and averaging the attention values within the overlapping areas. In our workflow, we employ Trident~\cite{zhang2025standardizing} to generate WSI-level attention visualizations (Figure~\ref{fig:diff_classes_attn_comparison} and Figure~\ref{fig:diff_dataset_attn_comparison}) and to display the patches with the highest attention scores. In Figure~\ref{fig:diff_models_attn_comparison}, UNI2 features generalize more robustly than Titan, and avoid artifact patches. We note that a high-attention score in CLL is given to non-lymphoma areas and in DLBCL to poorly preserved areas (Figure~\ref{fig:diff_classes_attn_comparison}). This behavior might be avoidable when training the algorithm with more classes. 

\section{Limitations}
Although this multicenter benchmark is comprehensive, several limitations remain. First, the substantial decrease in performance ($20\%$) on external test sets highlights unresolved out-of-distribution generalization challenges. Scanner-specific color variation, staining differences, and shifts in patient cohorts across clinical sites continue to affect model robustness, as reflected by the performance drop on OOD cohorts. Second, some cohorts exhibit class imbalance, particularly for rare subtypes or limited-control classes, which may bias model training and lead to overconfident predictions for majority classes.

\section{Conclusion}
We present the first multicenter lymphoma benchmark comprising 999 HE-stained whole slide images across four common lymphoma subtypes (CLL, FL, MCL, DLBCL) and healthy control tissue from four German centers. Combined with an automated MIL pipeline, we systematically evaluated five state-of-the-art pathology foundation models and two aggregation architectures across multiple magnifications.

Current “all-task” benchmark do not include lymphoma subtyping, and TCGA offers only a single lymphoma subtype (DLBCL) with limited histological diversity. In contrast, our multicenter benchmark focuses specifically on lymphoma, covering four common subtypes and healthy controls to enable a focused evaluation of this clinically important but so far underrepresented disease. On in-distribution test sets, all five foundation models (H-optimus-1, H0-mini, Virchow2, UNI2, Titan) achieved comparable performance with balanced accuracies exceeding $80\%$, demonstrating that current pathology foundation models have reached a similar level of representation quality for lymphoma subtyping. Both attention-based (AB-MIL) and transformer-based (TransMIL) aggregators performed similarly, with AB-MIL being computationally more efficient. Our magnification study revealed that 40× resolution is sufficient for accurate classification, with no additional benefit from higher resolutions or cross-magnification aggregation strategies. 
Despite strong in-distribution performance, we observed substantial performance drops on out-of-distribution test sets (balanced accuracy around $60\%$), revealing significant generalization challenges. This degradation likely stems from scanner-specific color variations, site-specific staining protocols, and differences in patient populations across centers. 
In contrast to expensive and time-consuming IHC staining, flow cytometry, and molecular genetic tests, our approach demonstrates that HE-stained WSIs can accurately determine common lymphoma subtypes when evaluated within the same clinical environment. An AI-augmented pathology workflow could guide initial screening to prioritize cases requiring expert review, potentially reducing turnaround times and allowing more efficient use of specialized diagnostic tests. 

To advance lymphoma diagnostics toward clinical utility, larger multicenter studies with standardized protocols are essential. We only have German centers included, but larger European and worldwide initiatives are needed. Future work should also expand coverage to include rare lymphoma subtypes, incorporate stain normalization or domain adaptation techniques to improve cross-site generalization, and conduct prospective clinical validation studies. Integrating complementary modalities such as immunophenotyping data or clinical parameters may further enhance diagnostic accuracy. Ultimately, such systems could assist pathologists in maximizing diagnostic yield from HE-stained sections while minimizing the number of auxiliary tests required for accurate diagnosis.

\clearpage  
\midlacknowledgments{C.M. acknowledges funding from the European Research Council (ERC) under the European Union's Horizon 2020 research and innovation program (Grant Agreement No. 866411 \& 101113551 \& 101213822) and support from the Hightech Agenda Bayern. L.W., S.R., R.S., and W.K. acknowledge funding by the BMBF Project Federated Digital Lymphoma Pathology (FDLP) (FKZ: 01KD2209A \& 01KD2415B).}

\section*{Ethics statement}
All experiments are conducted in accordance with the Declaration of Helsinki. The Technical University Munich (TUM)  Ethics Committee approved the retrospective analysis of lymphoma data (Approval No. 79/20 S-KH).

\bibliography{main}

@article{campanella2023computational,
  title={Computational Pathology at Health System Scale--Self-Supervised Foundation Models from Three Billion Images},
  author={Campanella, Gabriele and Kwan, Ricky and Fluder, Eugene and Zeng, Jennifer and Stock, Aryeh and Veremis, Brandon and Polydorides, Alexandros D and Hedvat, Cyrus and Schoenfeld, Adam and Vanderbilt, Chad and others},
  journal={arXiv preprint arXiv:2310.07033},
  year={2023}
}

@article{bray2021ever,
  title={The ever-increasing importance of cancer as a leading cause of premature death worldwide},
  author={Bray, Freddie and Laversanne, Mathieu and Weiderpass, Elisabete and Soerjomataram, Isabelle},
  journal={Cancer},
  volume={127},
  number={16},
  pages={3029--3030},
  year={2021},
  publisher={Wiley Online Library}
}

@article{chen2024towards,
  title={Towards a general-purpose foundation model for computational pathology},
  author={Chen, Richard J and Ding, Tong and Lu, Ming Y and Williamson, Drew FK and Jaume, Guillaume and Song, Andrew H and Chen, Bowen and Zhang, Andrew and Shao, Daniel and Shaban, Muhammad and others},
  journal={Nature Medicine},
  volume={30},
  number={3},
  pages={850--862},
  year={2024},
  publisher={Nature Publishing Group US New York}
}

@article{zhang2025standardizing,
  title={Accelerating Data Processing and Benchmarking of AI Models for Pathology},
  author={Zhang, Andrew and Jaume, Guillaume and Vaidya, Anurag and Ding, Tong and Mahmood, Faisal},
  journal={arXiv preprint arXiv:2502.06750},
  year={2025}
}

@article{ghia2007chronic,
  title={Chronic lymphocytic leukemia},
  author={Ghia, Paolo and Ferreri, Andr{\'e}s JM and Caligaris-Cappio, Federico},
  journal={Critical reviews in oncology/hematology},
  volume={64},
  number={3},
  pages={234--246},
  year={2007},
  publisher={Elsevier}
}

@article{xerri2016heterogeneity,
  title={The heterogeneity of follicular lymphomas: from early development to transformation},
  author={Xerri, Luc and Dirnhofer, Stephan and Quintanilla-Martinez, Leticia and Sander, Birgitta and Chan, John KC and Campo, Elias and Swerdlow, Steven H and Ott, German},
  journal={Virchows Archiv},
  volume={468},
  number={2},
  pages={127--139},
  year={2016},
  publisher={Springer}
}

@article{schieber2018current,
  title={Current overview and treatment of mantle cell lymphoma},
  author={Schieber, Michael and Gordon, Leo I and Karmali, Reem},
  journal={F1000Research},
  volume={7},
  pages={F1000--Faculty},
  year={2018}
}

@article{morton2006lymphoma,
  title={Lymphoma incidence patterns by WHO subtype in the United States, 1992-2001},
  author={Morton, Lindsay M and Wang, Sophia S and Devesa, Susan S and Hartge, Patricia and Weisenburger, Dennis D and Linet, Martha S},
  journal={Blood},
  volume={107},
  number={1},
  pages={265--276},
  year={2006},
  publisher={American Society of Hematology}
}

@article{vrabac2021dlbcl,
  title={DLBCL-Morph: Morphological features computed using deep learning for an annotated digital DLBCL image set},
  author={Vrabac, Damir and Smit, Akshay and Rojansky, Rebecca and Natkunam, Yasodha and Advani, Ranjana H and Ng, Andrew Y and Fernandez-Pol, Sebastian and Rajpurkar, Pranav},
  journal={Scientific Data},
  volume={8},
  number={1},
  pages={135},
  year={2021},
  publisher={Nature Publishing Group UK London}
}

@article{2010autocls,
author = {Orlov, Nikita and Chen, Wayne and Eckley, David and Macura, Tomasz and Shamir, Lior and Jaffe, Elaine and Goldberg, Ilya},
year = {2010},
month = {07},
pages = {1003-13},
title = {Automatic Classification of Lymphoma Images With Transform-Based Global Features},
volume = {14},
journal = {IEEE transactions on information technology in biomedicine : a publication of the IEEE Engineering in Medicine and Biology Society},
doi = {10.1109/TITB.2010.2050695}
}

@article{vorontsov2024foundation-virchow1,
  title={A foundation model for clinical-grade computational pathology and rare cancers detection},
  author={Vorontsov, Eugene and Bozkurt, Alican and Casson, Adam and Shaikovski, George and Zelechowski, Michal and Severson, Kristen and Zimmermann, Eric and Hall, James and Tenenholtz, Neil and Fusi, Nicolo and others},
  journal={Nature medicine},
  volume={30},
  number={10},
  pages={2924--2935},
  year={2024},
  publisher={Nature Publishing Group US New York}
}

@inproceedings{ilse2018attention,
  title={Attention-based deep multiple instance learning},
  author={Ilse, Maximilian and Tomczak, Jakub M and Welling, Max},
  booktitle={International conference on machine learning},
  pages={2127--2136},
  year={2018},
  organization={PMLR}
}

@article{shao2021transmil,
  title={Transmil: Transformer based correlated multiple instance learning for whole slide image classification},
  author={Shao, Zhuchen and Bian, Hao and Chen, Yang and Wang, Yifeng and Zhang, Jian and Ji, Xiangyang and others},
  journal={Advances in Neural Information Processing Systems},
  volume={34},
  pages={2136--2147},
  year={2021}
}

@article{chen2024uni,
  title={Towards a General-Purpose Foundation Model for Computational Pathology},
  author={Chen, Richard J and Ding, Tong and Lu, Ming Y and Williamson, Drew FK and Jaume, Guillaume and Chen, Bowen and Zhang, Andrew and Shao, Daniel and Song, Andrew H and Shaban, Muhammad and others},
  journal={Nature Medicine},
  publisher={Nature Publishing Group},
  year={2024}
}

@inproceedings{sens2023bel,
  title={BEL: A bag embedding loss for transformer enhances multiple instance whole slide image classification},
  author={Sens, Daniel and Sadafi, Ario and Casale, Francesco Paolo and Navab, Nassir and Marr, Carsten},
  booktitle={2023 IEEE 20th International Symposium on Biomedical Imaging (ISBI)},
  pages={1--5},
  year={2023},
  organization={IEEE}
}

@software{hoptimus1,
  author = {Bioptimus},
  title = {H-optimus-1},
  url = {https://huggingface.co/bioptimus/H-optimus-1},
  year = {2025},
}

@article{zimmermann2024virchow2,
  title={Virchow2: Scaling Self-Supervised Mixed Magnification Models in Pathology}, 
  author={Eric Zimmermann and Eugene Vorontsov and Julian Viret and Adam Casson and Michal Zelechowski and George Shaikovski and Neil Tenenholtz and James Hall and Thomas Fuchs and Nicolo Fusi and Siqi Liu and Kristen Severson},
  journal={arXiv preprint arXiv:2408.00738},
  year={2024},
}

@inproceedings{h0-mini,
  title={Distilling foundation models for robust and efficient models in digital pathology},
  author={Filiot, Alexandre and Dop, Nicolas and Tchita, Oussama and Riou, Auriane and Dubois, R{\'e}my and Peeters, Thomas and Valter, Daria and Scalbert, Marin and Saillard, Charlie and Robin, Genevi{\`e}ve and others},
  booktitle={International Conference on Medical Image Computing and Computer-Assisted Intervention},
  pages={162--172},
  year={2025},
  organization={Springer}
}

@article{ding2025titan,
  title={A multimodal whole-slide foundation model for pathology},
  author={Ding, Tong and Wagner, Sophia J and Song, Andrew H and Chen, Richard J and Lu, Ming Y and Zhang, Andrew and Vaidya, Anurag J and Jaume, Guillaume and Shaban, Muhammad and Kim, Ahrong and others},
  journal={Nature Medicine},
  pages={1--13},
  year={2025},
  publisher={Nature Publishing Group US New York}
}

@inproceedings{he2016resnet50,
  title={Deep residual learning for image recognition},
  author={He, Kaiming and Zhang, Xiangyu and Ren, Shaoqing and Sun, Jian},
  booktitle={Proceedings of the IEEE conference on computer vision and pattern recognition},
  pages={770--778},
  year={2016}
}

@article{campanella2019clinical,
  title={Clinical-grade computational pathology using weakly supervised deep learning on whole slide images},
  author={Campanella, Gabriele and Hanna, Matthew G and Geneslaw, Luke and Miraflor, Allen and Werneck Krauss Silva, Vitor and Busam, Klaus J and Brogi, Edi and Reuter, Victor E and Klimstra, David S and Fuchs, Thomas J},
  journal={Nature medicine},
  volume={25},
  number={8},
  pages={1301--1309},
  year={2019},
  publisher={Nature Publishing Group US New York}
}

@inproceedings{sadafi2020attention,
  title={Attention based multiple instance learning for classification of blood cell disorders},
  author={Sadafi, Ario and Makhro, Asya and Bogdanova, Anna and Navab, Nassir and Peng, Tingying and Albarqouni, Shadi and Marr, Carsten},
  booktitle={International Conference on Medical Image Computing and Computer-Assisted Intervention},
  pages={246--256},
  year={2020},
  organization={Springer}
}

@article{bareja2025evaluating,
  title={Evaluating Vision and Pathology Foundation Models for Computational Pathology: A Comprehensive Benchmark Study},
  author={Bareja, Rohan and Carrillo-Perez, Francisco and Zheng, Yuanning and Pizurica, Marija and Nandi, Tarak Nath and Shen, Jeanne and Madduri, Ravi and Gevaert, Olivier},
  journal={medRxiv},
  pages={2025--05},
  year={2025},
  publisher={Cold Spring Harbor Laboratory Press}
}

@article{khoury20225th,
  title={The 5th edition of the World Health Organization classification of haematolymphoid tumours: myeloid and histiocytic/dendritic neoplasms},
  author={Khoury, Joseph D and Solary, Eric and Abla, Oussama and Akkari, Yassmine and Alaggio, Rita and Apperley, Jane F and Bejar, Rafael and Berti, Emilio and Busque, Lambert and Chan, John KC and others},
  journal={leukemia},
  volume={36},
  number={7},
  pages={1703--1719},
  year={2022},
  publisher={Nature Publishing Group UK London}
}

@inproceedings{ozlugedik2026moa,
  title={MoA: Mixture of Aggregators Improves Slide-Level Diagnosis in Computational Pathology},
  author={Ozlugedik, Fatih Ibrahim and Dasdelen, Muhammed Furkan and Umer, Rao Muhammad and Marr, Carsten},
  booktitle={Medical Imaging with Deep Learning},
  year={2026}
}

\end{document}